\documentclass[a4paper]{article}
\usepackage[english]{babel}
\usepackage[utf8x]{inputenc}
\usepackage[T1]{fontenc}
\usepackage[]{algorithm2e}
\usepackage{indentfirst}
\usepackage{authblk}
\usepackage[a4paper,top=3cm,bottom=2cm,left=3cm,right=3cm,marginparwidth=1.75cm]{geometry}
\usepackage{amsmath}
\usepackage{graphicx}
\usepackage[colorinlistoftodos]{todonotes}
\usepackage[colorlinks=true, allcolors=blue]{hyperref}

\title{Gradient target propagation}

\author{Tiago de Souza Farias\footnote{tiago939@gmail.com}\mbox{ }}
\author{Jonas Maziero\footnote{jonas.maziero@ufsm.br}}
\affil{Departamento de F\'isica, Centro de Ci\^encias Naturais e Exatas, Universidade Federal de Santa Maria, Avenida Roraima 1000, Santa Maria, RS, 97105-900, Brazil}
\date{}

\begin{document}
\maketitle

\begin{abstract}
We report a learning rule for neural networks that computes how much each neuron should contribute to minimize a giving cost function via the estimation of its target value. By theoretical analysis, we show that this learning rule contains backpropagation, Hebbian learning, and additional terms.  We also give a general technique for weights initialization. Our results are at least as good as those obtained with backpropagation. The neural networks are trained and tested in three problems: MNIST, Fashion-MNIST, and CIFAR-10 datasets. The associated code is available at \url{https://github.com/tiago939/target}.
\end{abstract}
\section{Introduction}

The search for an efficient learning rule for artificial neural networks is an active area of research. A learning rule is a set of equations that controls how much the tunable parameters of a neural network must change in order for an objective to be accomplished. Usually, these parameters are the synaptic strengths between neurons, called weights. But others parameters, such as bias, activation function or even the learning rules themselves, can be considered \cite{8}.

\begin{figure}[h]
\begin{center}
\includegraphics[width=0.5\textwidth]{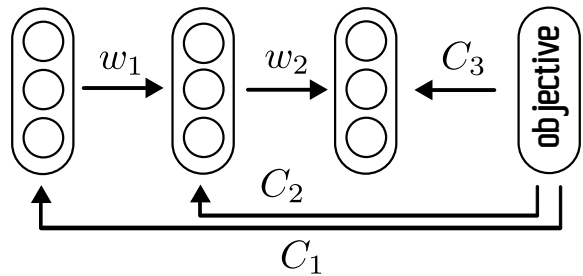}
\end{center}
\caption{This image illustrates the backpropagation channels. The feedbacks $C_{j}$ are computed from the objective to each layer. The numbers $w_1$ and $w_2$ are the weights, which are the learning parameters here.}
\label{fig:backprop}
\end{figure}

Pure local learning rules (i.e., rules in which the parameters changes depend solely on local information between the pre-synaptic and post-synaptic neurons), such as Hebbian learning \cite{4}, are insufficient for optimal learning, as shown in \cite{8}. An efficient learning rule must provide feedback from non-local information in order to guide the neural network towards an objective. The source of feedback comes from the definition of the objective, such as the error function, goal function, reward or mutual information \cite{dl}. Channels between non-neighbor neurons provide the necessary learning feedback, as shown in Fig. \ref{fig:backprop}.

The credit assignment problem relates how much each neuron should contribute to the network for an objective to be reached. The amount of neuron credit is called target. If feedback is provided, a neuron will have a target value, thus enabling learning.

In most neural networks architectures and learning frameworks, external feedback is not provided to the whole network, but only to a group of neurons. This group of neurons is related to the actions of the network that controls the distance to the objective.

If the feedback is incomplete,  there are no targets for the whole network, prohibiting training every neuron. One popular method to get around this problem is to use the backpropagation method ~\cite{3}: instead of worrying about targets, backpropagation concerns directly with the learning parameters, by using the derivatives of error equipped with the chain-rule.

Despite the popularity of the backpropagation algorithm, it has undesired problems, such as locking problem, vanishing or exploding gradients, continuous and unsupervised learning ~\cite{13}. Some of them are solved with tricks, such as batch normalization ~\cite{11} and dropout ~\cite{22}. Still, a learning rule that could avoid some, if not all, of these problems would be desirable.

Besides the problem with the learning rule, the random distribution of parameters for initialization of a neural network can be a reason for great concern ~\cite{12}. At the beginning of training, the initial learning parameters must be chosen. A neural network can learn very slowly or not learn at all if a bad distribution is used.

The purpose of this paper is twofold. First, we show a learning rule that calculates and propagates targets for all neurons. Secondly, we derive a technique for weight initialization for optimal learning.

The remainder of this article is organized as follows. In Sec. \ref{sec:rw} we discuss related work for both target propagation and parameter distribution. In Sec. \ref{sec:methods}, we present the method of gradient target propagation (Sec. \ref{sec:target}) and our approach for weight initialization (Sec. \ref{sec:weight}). In Sec. \ref{sec:results} we report the results obtained for three different problems: MNIST (Sec. \ref{sec:mnist}), Fashion-MNIST (Sec. \ref{sec:mnistfashion}), and CIFAR-10 datasets (Sec. \ref{sec:cifar10}). In Sec. \ref{sec:disc} we discuss our results and in Sec. \ref{sec:conc} we make some final remarks.

\section{Related work}
\label{sec:rw}
The credit assignment problem has already been studied in the literature, and some techniques were developed to calculate the target values of neurons \cite{5,7}. As will be discussed in Sec. \ref{sec:methods}, targets cannot be computed directly, since it is not possible to invert the function with respect to the desired activation without avoiding coupled equations (with exception of some very restricted conditions, useless for most problems of interest).

Bengio proposed a method \cite{6}, where the targets are computed by an approximated inverse of the non-linear function. This approximation is learned by the neural network in an auto-encoded fashion. Results comparable to the backpropagation method were obtained for MNIST and CIFAR-10 data sets. The drawbacks of this approach are that the inverse must be learned and that it is only approximated and is limited to the space of the auto-encoder.

Target sampling is another method already considered in the literature \cite{8}. In this method, the target values are drawn from a random distribution. For a set of samples, each one is evaluated by an objective function and the sample with the best measurement is used to update the learning parameters. While this technique can work for small networks, the sampling space grows exponentially with the number of neurons, what makes this method computationally inefficient.

Local Representation Alignment \cite{23} is an alternative approach to calculate the target values, by computing them with gradient steps. There are two targets considered: the credit of the input to each neuron and the credit of the output. A normalization constraint is required to ensure the targets are within the reach of neural activity.

One may wonder if merging a set of rules could accelerate the learning of a neural network. A specific combination of two rules: backpropagation and Hebbian learning was done in ~\cite{2}. While it was found that this hybrid rule is no better than backpropagation, the results do not exclude the possibility for better performance on other set of problems or that other combinations could improve performance.

Weight initialization is a problem well known to the machine learning community. Xavier's initialization ~\cite{10} is a common approach to mitigate the effects of bad weights. This initialization relies on tools from statistics, and assumes a random distribution of weights with  variance inversely proportional to the number of neurons in each layer. However, this method relies on the assumption of linear activation functions, which is unrealistic.

A better approach was reported in ~\cite{1}, which contains the Xavier initialization as a special case. There the non-linearity of the activation function was considered, so that for each function we will have a different random distribution of parameters. The problem with this approach is that it assumes a specific distribution for the input data.

\section{Methods}
\label{sec:methods}

Here we consider a feed-forward neural network with $L$ layers. We denote $y_l^{(i)}$ the activation of the $i$-th neuron from the $l$-th layer. The target value for this neuron activation shall be denoted with a hat: $\hat{y_l}^{(i)}$. If the super-index is not present, the corresponding symbol represents the activation vector for all neurons of layer $l$.

The activation $y$ is a non-linear function $f$ of its inputs, which are given as a linear combination of the product of the weights and activations from the previous layer:
\begin{equation}
y_l = f(z_l)\textrm{, with } z_l=\sum_{j} w_l^{(j)} y_{l-1}^{(j)}.
\end{equation}
In this work we use the sigmoid activation function:
\begin{equation}
f(z) = \frac{1}{1+e^{-z}}.
\end{equation}
\label{eq:target}
\subsection{Gradient targets}
\label{sec:target}

In the supervised learning framework, the data to train a neural network is labeled. These labels control how much activation each neuron from the output layer should have, and these values are the targets $\hat{y}_L$.

Let $C$ be the cost function to be minimized. The target $\hat{y}_l$ can be obtained by solving the autonomous differential equation:
\begin{equation}
\frac{\partial y_l}{\partial t} = -\frac{\partial C_{l+1}}{\partial y_l},
\label{eq:target}
\end{equation}
with $C_{l+1}=\sum_{i}C_{l+1}^{(i)}$ and 
each neuron has its own local cost function, which must depend on the target and actual activation. One possible choice is the quadratic cost function:
\begin{equation}
C_{l+1}^{(i)} = \frac{1}{2} \left(\hat{y}_{l+1}^{(i)} - y_{l+1}^{(i)}\right)^2.
\label{eq:costL}
\end{equation}

The true target of $y_l$ is the solution of (\ref{eq:target}) at infinite time: $\hat{y}_l = y_l(t)|_{t=\infty}$. The only way to know the true targets is to solve the equation analytically. However, this is unfeasible for layers with more than one neuron (see Appendix C). Nevertheless, we can obtain an approximation to it by solving the equation numerically using a truncated time T. As the true targets can lead to lower cost, so does the approximated targets, since they are closer to the optimal network than the actual activation of the neurons.

\begin{figure}[h]
\begin{center}
\includegraphics[width=0.5\textwidth]{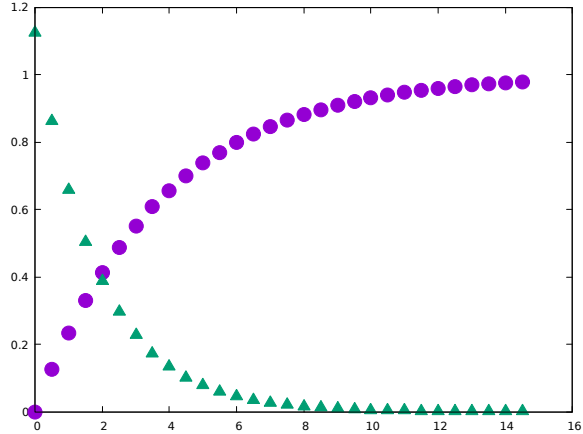}
\end{center}
\caption{Time (x-axis) evolution of the target (y-axis) computed by eq. (\ref{eq:wupd}) for a simple case of two neurons network. Circle: target value, triangle: square error. As the time is increased, the target value converges to a value where the error is minimal.}
\label{fig:time}
\end{figure}

From a numerical point of view, the time step size and period can affect the convergence of the targets. One option is to treat them as hyper-parameters. Then it becomes a question of finding their optimal values. More technical details about numerical methods are given in Appendix B.

By its turn, the weights are updated using the gradient descent method:
\begin{equation}
\Delta w_l = -\eta \frac{\partial C_{l}}{\partial w_{l}},
\label{eq:wupd}
\end{equation}
where $\eta$ is the learning rate hyper-parameter. An example appears in figure \ref{fig:time}.

Note that in order to get the targets, first the output cost function must be computed. The targets are then computed from the output layer to the input layer, backwards with respect to information direction of the neural network. Thus, the pre-synaptic neuron sends the information signal to its post-synaptic neurons, and they send back the target value.

\subsection{Weight initialization}
\label{sec:weight}

To avoid bad distributions of random parameters, we derive a statistical technique for the initialization of weights. We followed steps similar to ~\cite{1}, but with more generality.

As shown in the Appendix A, we obtained the following variance for the weights normal distribution:
\begin{equation}
VAR(w_l) = \frac{VAR(y_l)}{N f'(\langle z_{l} \rangle)^2 (\langle y_{l-1} \rangle + VAR(y_{l-1}))},
\end{equation}
which is more general than the weight initialization mentioned in Section \ref{sec:rw}.

\begin{figure}[h]
\begin{center}
\includegraphics[width=0.5\textwidth]{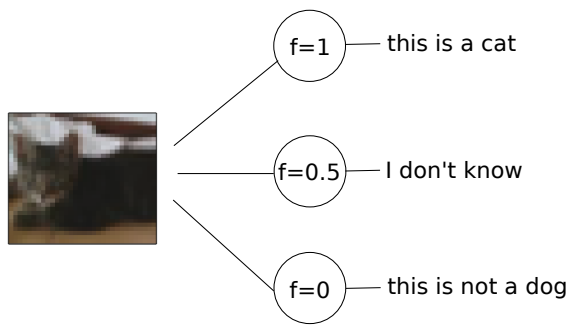}
\end{center}
\caption{The uncertainty of a neuron is related with its activation value. Image from the CIFAR-10 dataset.}
\label{fig:catnn}
\end{figure}

The problem consists in determining the variance of the activation function. This is solved by defining the amount of knowledge each neuron has about the data. Neuron uncertainty characterizes how much unsure a neuron is about its input values. This uncertainty is quantified by the activation value of the neuron: if the activation is very high or very low, the uncertainty is low. For example, say a neuron is trained to identify if an image has a cat or a dog in it. If this neuron has a high (low) activation, then it means this neuron is sure this image (not) has a cat (see figure \ref{fig:catnn}). The sigmoid function has values in the range (0,1). The extreme values corresponds to high confidence and the middle value (1/2) to low confidence.

In the beginning of training, for a  neural network that never received the data before, we would like to have the neurons with as much uncertainty as possible. One can obtain this by choosing a distribution of weights in which this uncertainty is maximized. For every activation function there are an infinite possible uncertainty functions we could use. For the sigmoid function, one example is the following:
\begin{equation}
U(y) = 6y(1-y).
\end{equation}
The factor 6 comes from the probability interpretation of $U$. Since the weights are draw randomly, we relate $U(y)$ to the probability of uncertainty:
\begin{equation}
\int_0^1 U(y) = 1.
\end{equation}

As shown in Appendix A, with the condition of sigmoid neurons (with exception to the input layer) and normalized data, the distribution of random weights must be normal with mean zero and variance
\begin{equation}
VAR(w_1) = \frac{48}{35 N_1}
\end{equation}
for the input layer to the first hidden layer and
\begin{equation}
VAR(w_l) = \frac{16}{11 N_l}
\end{equation}
for all the other layers.

The algorithm \ref{alg:algtarget} summarizes the technique of gradient target propagation:

\begin{algorithm*}[H]
 Initialize the neural network with the weights distribution as discussed above and define the hyper-parameters\;
\For{$l=N$ to $1$}{
 Calculate the $l$-th error $C_l$\;
 Calculate the ($l-1$)-th target by $\hat{y}_t = \hat{y}_{t-1} - \tau \frac{\partial C_l}{\partial y_{l-1}}$\;
 Calculate the weight change from the ($l-1$)-th to $l$-th layer by $\Delta w_{l} = -\eta \frac{\partial C_l}{\partial w_{l}}$\;
 }
 Update the weights $w_{l} = w_{l} + \Delta w_{l}$
\caption{Gradient target propagation algorithm}
\label{alg:algtarget}
\end{algorithm*}

\subsection{Target propagation as a general rule}
\label{sec:targen}

The target propagation can be shown to be a more general method of learning than pure local or non-local rules. In fact, here we will show that what it actually does is a combination of both.

From the weight change we have:
\begin{equation}
\Delta w_l = -\eta \frac{\partial C_{l}}{\partial w_l}
\end{equation}
Now we consider the cost with the squared error:
\begin{equation}
\Delta w_l = \eta (\hat{y}_{l} - y_{l}) \frac{\partial y_{l}}{\partial w_{l}}
\end{equation}
Following we can separate the equation above:
\begin{equation}
\Delta w_l = \eta \hat{y}_{l} \frac{\partial y_{l}}{\partial w_{l}} - \eta y_l \frac{\partial y_{l}}{\partial w_{l}}
\end{equation}

Replacing the target with the equation (\ref{eq:target}) for one instant of time:
\begin{equation}
\Delta w_l = -\eta \tau \frac{\partial C_{l+1}}{\partial y_{l}} \frac{\partial y_{l}}{\partial w_{l}} - \eta y_{l} \frac{\partial y_{l}}{\partial w_{l}}
\end{equation}

The first term is just the backpropagation chain-rule. So this technique is doing backpropagation plus something else. The second term can be rewritten as the derivative of the non-linear function f:
\begin{equation*}
y_{l} \frac{\partial y_{l}}{\partial w_{l}} = y_{l-1} y_l \frac{\partial f(z_{l})}{\partial z_l}
\end{equation*}
This corresponds to the non-linear Hebbian-like learning ~\cite{14}, which is a local learning rule.

The cost in the equations above can be rewritten in terms of the cost from layers above, thus generalized for all forms. For one single time step, one can find that (see appendix D):
\begin{equation}
\Delta w_{l} = -\frac{\partial}{\partial w_{l}} \left(C_{q} + \sum_{i=l}^{q-1} \frac{y_i^2}{2} \right)
\end{equation}

In the continuous form, we have:

\begin{equation}
\Delta w_{l} = -\frac{\partial}{\partial w_l} \left( \int_0^T C_q d^{q-l} t + \sum_{i=l}^{q-1} \int_0^T \frac{y_i^2}{2} d^{i-l}t \right)
\end{equation}

\begin{figure}[h]
\begin{center}
\includegraphics[width=0.5\textwidth]{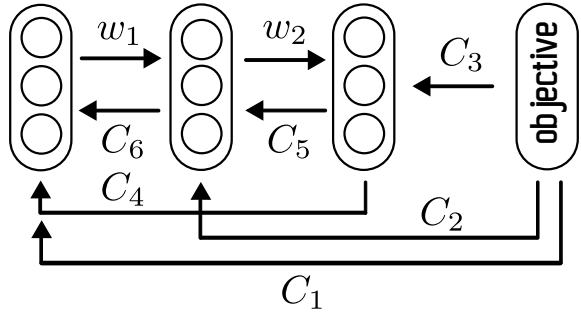}
\end{center}
\caption{Target propagation channels, feedback $C$ is computed from different sources. $C_1$, $C_2$ and $C_3$ come from back-propagation. $C_5$ and $C_6$ have local Hebbian nature. $C_4$ is a non-local channel outside the objective function. }
\label{fig:targetprop}
\end{figure}

From the equations above, it is easy to see that target propagation combines non-local learning rules and local learning rules. By adding more feedback channels (figure \ref{fig:targetprop}), the targets become possible to be calculated.

\section{Results}
\label{sec:results}

Unless specified otherwise, the neurons of the first layer has ReLU activation function and all neurons from the other layers are sigmoid. All networks are feed-forward.

Each dataset is trained for a number of epochs and the hyper-parameters were adjusted empirically. All data was normalized to the range [0,1], by dividing each point by the maximal value of the set. The mini-batch size is one: the weights are updated after each example in the training data set.

The labels feed into the network are set as one-hot encoding. In this manner, the targets for the output layer are:
\begin{table}[h]
\centering
\begin{tabular}{ l|c }
 class  & target vector  \\
 class 1  &  (1,0,...,0) \\
 class 2  &  (0,1,...,0) \\
 ...  &  ... \\
 class N  &  (0,0,...,1) \\ & 
\end{tabular}
\end{table}

The neural networks were developed in the Python programming language. The autonomous differential equation was solved with the Euler method (see the appendix C for technical details).

The 1-hidden and 2-hidden layers size networks were trained in 30 epochs. The 3-hidden layer network were trained in 60 epochs.

\subsection{MNIST}
\label{sec:mnist}

The MNIST~\cite{15} is a dataset of 70000 handwritten numbers from zero to nine. Each image has a size of 28x28 pixels in gray scale. As common practice, we divided the dataset into 50000 images for training and 10000 images for testing.  Table \ref{table:mnist} show the results on the testing set.

\begin{table}[ht]
\centering
\begin{tabular}{ l | c | r | r }
 Network  & Score  & $\tau$ & $\eta$ \\
 784-100-10  &  97.21\% & 400 & 0.01 \\
 784-200-10 &  97.5\% & 400 & 0.01 \\
 784-500-10 &  97.81\% & 400 & 0.01 \\
 784-100-100-10 & 97.31\% & 50 & 0.001 \\
 784-500-500-10 & 98.23\% & 50 & 0.001 \\
 784-500-500-500-10 & 98.23\% & 25 & 0.0005 \\
\end{tabular}
\caption{Results on MNIST}
\label{table:mnist}
\end{table}

\subsection{Fashion-MNIST}
\label{sec:mnistfashion}

The Fashion-MNIST~\cite{16} is a dataset that contains 70000 images of 10 different classes of clothing. The images have size 28x28 pixels in gray-scale. Just as the MNIST problem, we divided the dataset in 50000 images for training and 10000 images for testing.

\begin{table}[ht]
\centering
\begin{tabular}{ l | c | r | r }
 Network  & Score  & $\tau$ & $\eta$ \\
 784-100-10  &  87.03\% & 100 & 0.01 \\
 784-500-10 &  87.29\% & 100 & 0.01\\
 784-100-100-10  & 87.71\% & 25 & 0.0005\\
 784-500-500-10  & 87.77\% & 25 & 0.0005\\
 784-100-100-100-10 & 87.92 \% & 25 & 0.000025 \\
\end{tabular}
\caption{Results on Fashion-MNIST}
\end{table}

\subsection{CIFAR-10}
\label{sec:cifar10}

The CIFAR-10 dataset~\cite{17} consists in 60000 images of 10 different classes, from cats to trucks. Each image has size 32x32 pixels in RGB color channels.

Since each image has three colors, the input layer has 32x32x3 neurons.

\begin{table}[ht]
\centering
\begin{tabular}{ l | c | r | r }
 Network  & Score  & $\tau$ & $\eta$ \\
 3072-100-10  &  42.30\% & 50 & 0.001 \\
 3072-500-10 &   44.74\% & 50 & 0.001\\
 3072-100-100-10 & 44.89\% & 50 & 0.0001\\
 3072-500-500-10 & 47.06\% & 50 & 0.0001 \\
 3072-500-500-500-10 & 47.11 \% & 50 & 0.000005\\
\end{tabular}
\caption{Results on CIFAR-10}
\end{table}


\section{Discussion}
\label{sec:disc}

The results on performance with gradient target propagation are comparable~\cite{15, 16, 18} to other learning rules for all data sets. This technique is able to train neural networks of arbitrary size and data, limited only by the computational resources available. The implementation of gradient target propagation is accessible, although computing the target values is restricted to the numerical method of choice, the number of time steps and the step size are what really put bounding constrains to stability.

While the performance on CIFAR-10 is low, this issue is related to the neural network architecture (the ones trained here are feed-forward) rather than the technique itself, since backpropagation also leads to the same order of performance. Target propagation can be applied to different architectures, like convolutional and recurrent neural networks, and might lead to better results.

The choice of hyper-parameters also limits the learning performance of neural networks. In this work we chose one learning rate and one time step for all neurons, which degrades the training time as the number of layers is increased (higher number of hidden layers requires lower learning rates because the approximated targets get bigger as the network grows). A better approach is to set different hyper-parameters to different layers or, ideally, one unique for each neuron, which can be provided by alternative optimization techniques, such as accelerate methods~\cite{19}, genetic algorithms~\cite{20}, and Bayesian optimization~\cite{21}. 

The equation (6) suggest problems for activation functions in which the derivative at zero is zero, then we would get infinite variance. This comes fundamentally from the expected value of the weights, which was set to zero in this work. However, most of the non-linear functions have non zero value at the derivative (even the ReLU function, which can be set to one at this point). Nevertheless, if this problem is unavoidable we can choose to either make an approximation or set the expected value to be centered to other numbers, then the distribution will be reshaped to meet the new criteria and maintain the uncertainty distribution of activation at maximum, as desired. 


\section{Conclusion}
\label{sec:conc}

We presented an alternative to backpropagation to train neural networks that has lead to results competitive with other techniques. By assigning each neuron with a target value, we showed that artificial neural networks can be trained. We also reported a technique to distribute the initial random weights of a neural network by using the idea of neuron uncertainty. 

Theoretical analysis on the target equations lead to the important observation that target propagation combines a set of learning rules which depend on local information, including backpropagation and Hebbian learning. The additional, non-local, learning rules provide the necessary information to determine how much each neuron should contribute to achieve a desired goal.

We have provided the numerical tools for implementation of gradient target propagation and we also gave the mathematical background for its understanding. This technique can be explored a lot further in the future, by both numerical and theoretical points of view, since here it was applied in its most simplified form.


\section*{Acknowledgments}
This work was supported by the Brazilian National Institute for the Science and Technology of Quantum Information (INCT-IQ), process 465469/2014-0, and by the Coordination for the Improvement of Higher Education Personnel (CAPES).

\bibliographystyle{ieeetr}
\bibliography{sample}

\section*{Appendix A}
\label{appendix:app1}
We start with a general non-linear function which depends on the sum of the weights and inputs as folows:
\begin{equation}
y_l = f(z_l) \text{, with} \quad z_l=\left( \sum_i^N w^{(i)}_l y_{l-1}^{(i)} \right).
\end{equation}

The variance of $y_l$ can be computed with the approximation:
\begin{equation}
VAR(y_l) \approx \left( \frac{df}{dz_l} (\langle z_l \rangle) \right)^2 VAR (z_l).
\label{eq:var1}
\end{equation}

Since $w^{(i)}_{l}$ and $y_{l-1}^{(i)}$ are independent, the expected value of $z_l$ can be written as
\begin{equation}
\langle z_l \rangle = \langle \sum_i w^{(i)}_{l} y_{l-1}^{(i)} \rangle = \sum_i \langle w^{(i)}_{l} \rangle \langle y^{(i)}_{l-1} \rangle
\end{equation}

Since each $w^{(i)}_{l}$ comes from the same distribution, all the expected values are the same: $\langle w^{(i)}_{l} \rangle = \langle w_l \rangle $. If we choose the normal distribution centered at zero for the weights, then:
\begin{equation}
\langle z_l \rangle = \langle w_l \rangle \sum_i \langle y_{l-1}^{(i)} \rangle = 0
\end{equation}

Now we need the variance of $z_l$:
\begin{align}
VAR(z_l) & = VAR \left( \sum_i w^{(i)}_{l} y^{(i)}_{l-1} \right) \\
& = \sum_i VAR (w^{(i)}_{l} y^{(i)}_{l-1}) \\
& = \sum_i \langle w^{(i)}_{l} \rangle ^2 VAR(y^{(i)}_{l-1}) + \langle y^{(i)}_{l-1} \rangle ^2 VAR(w^{(i)}_{l}) + VAR(w^{(i)}_{l})VAR(y^{(i)}_{l-1}) \\
& = \sum_i \langle y^{(i)}_{l-1} \rangle ^2 VAR(w^{(i)}_{l}) + VAR(w^{(i)}_{l})VAR(y^{(i)}_{l-1}) \\
& = VAR(w_l) \left( \sum_i \langle y^{(i)}_{l-1} \rangle ^2 + VAR(w^{(i)}_{l})VAR(y^{(i)}_{l-1}) \right)
\end{align}

If we assume all inputs have the same expected value and variance, then:
\begin{equation}
VAR(z_l) = VAR(w_l) N (\langle y_{l-1} \rangle + VAR(y_{l-1})
\end{equation}

Thus, from equation (\ref{eq:var1}):
\begin{equation}
VAR(w_l) = \frac{VAR(y_l)}{N f'(0)^2 (\langle y_{l-1} \rangle + VAR(y_{l-1}))}
\end{equation}

From the concept of neuron uncertainty, we have:
\begin{align}
& \langle y \rangle = \int_0^1 y 6y(1-y) dy = \frac{1}{2},\\
& VAR(y) = \int_0^1 (y-\langle y \rangle)^2 6y(1-y) dy = \frac{1}{20}.
\end{align}

For the input layer, we can assume a distribution for the data or evaluate it. Here we assume a uniform distribution for the input data, which was normalized from 0 to 1. This uniform distribution has expected value $\langle x \rangle = 0.5$ and variance $1/12$. Thus, the weight distribution for the input layer has the form:
\begin{equation}
VAR(w_1) = \frac{48}{35 N_1}.
\end{equation}
And, since we have the variance and expected value for all the other layers:
\begin{equation}
VAR(w_l) = \frac{16}{11 N_l}.
\end{equation}

\section*{Appendix B: Euler method}
\label{appendix:app2}
The autonomous differential equation can be solved with the Euler method, for each time t:
\begin{equation}
y^t_l = y^{t-1}_l - \tau \frac{\partial C_{l+1}}{\partial y_l},
\end{equation}
where $\tau$ is the time step size and the equation is solved with $T$ iterations, in which the target of $y$ is determined by $\hat{y} = y^T$.

There is a trade-off between the number of iterations $T$ and the step size. The greater the step size, the fewer iterations are needed, but less accurate is from the true target. However, numerical instabilities must be considered. Here we treated them as hyper-parameters, where they can depend on the layer or not.

\section*{Appendix C: Proof that targets are not invertible}
\label{appendix:app3}

We begin with the linear sum of weights and inputs. Denote $w^{(j:i)}$ the weights from the $j$-th neuron to the $i$-th neuron.
\begin{equation}
y_{l+1}^{(i)} = f\left( \sum_j w^{(j:i)}_l y^{(j)}_{l} \right).
\end{equation}
Let us assume now that $f$ is invertible:
\begin{equation}
f^{-1} (y_{l+1}^{(i)}) = \sum_j w^{(j:i)}_l y^{(j)}_{l}
\end{equation}

Now we choose a neuron $k$:
\begin{equation}
y^{(k)}_l = \frac{1}{w^{(k:i)}_l} \left( f^{-1} (y_{l+1}^{(i)}) - \sum_{j\neq k} w^{(j:i)}_l y^{(j)}_{l} \right)
\end{equation}

Then we sum over all post-synaptic neurons:
\begin{equation}
y^{(k)}_l = \frac{1}{N} \sum_i^N \frac{1}{w^{(k:i)}_l} \left( f^{-1} (y_{l+1}^{(i)}) - \sum_{j\neq k} w^{(j:i)}_l y^{(j)}_{l} \right)
\end{equation}

So, the targets to layer $l$ are:
\begin{equation}
\hat{y}_l^{(k)} = \frac{1}{N} \sum_i^N \frac{1}{w^{(k:i)}_l} \left( f^{-1} (\hat{y}_{l+1}^{(i)}) - \sum_{j\neq k} w^{(j:i)}_l \hat{y}^{(j)}_{l} \right)
\end{equation}

Since all targets are evaluated at the same time and $y^{(k)}_l$ depends on all the other targets from the same layer, it is not possible to find an analytical solution to the autonomous differential equation, unless this layer has only a single neuron.

\section*{Appendix D: Target propagation as a general learning rule}
\label{appendix:app4}
We start from the quadratic cost:
\begin{equation}
C_l = \frac{1}{2} (\hat{y_l} - y_l)^2.
\end{equation}
Omitting the hyper-parameters $\tau$ and $\eta$, for a single time step:
\begin{equation}
\hat{y}_{l} = -\frac{\partial C_{l+1}}{\partial y_{l}}.
\end{equation}
So,
\begin{align}
\Delta w_{l} & = - \frac{\partial C_{l}}{\partial w_{l}} \\
& = (\hat{y}_{l} - y_{l}) \frac{\partial y_{l}}{\partial w_{l}} \\
& = \left( -\frac{\partial C_{l+1}}{\partial y_{l}} - y_{l} \right) \frac{\partial  y_{l}}{\partial w_{l}} \\
& = -\frac{\partial C_{l+1}}{\partial y_{l}} \frac{\partial y_{l}}{\partial w_{l}} - y_{l} \frac{\partial y_{l}}{\partial w_{l}} \\
& = (\hat{y}_{l+1} - y_{l+1}) \frac{\partial y_{l+1}}{\partial y_{l}} \frac{\partial y_{l}}{\partial w_{l}} - y_{l} \frac{\partial y_{l}}{\partial w_{l}} \\
& = -\frac{\partial C_{l+2}}{\partial y_{l+1}} \frac{\partial y_{l+1}}{\partial y_{l}} \frac{\partial y_{l}}{\partial w_{l}} - y_{l+1} \frac{\partial y_{l+1}}{\partial y_{l}} \frac{\partial y_{l}}{\partial w_{l}} - y_{l} \frac{\partial y_{l}}{\partial w_{l}} \\
& = (\hat{y}_{l+2} - y_{l+2})\frac{\partial y_{l+2}}{\partial y_{l+1}}\frac{\partial y_{l+1}}{\partial y_{l}} \frac{\partial y_{l}}{\partial w_{l}} - y_{l+1} \frac{\partial y_{l+1}}{\partial y_{l}} \frac{\partial y_{l}}{\partial w_{l}} - y_{l} \frac{\partial y_{l}}{\partial w_{l}} \\
& = \left(-\frac{\partial C_{l+3}}{\partial y_{l+2}} - y_{l+2}\right)\frac{\partial y_{l+2}}{\partial y_{l+1}}\frac{\partial y_{l+1}}{\partial y_{l}} \frac{\partial y_{l}}{\partial w_{l}} - y_{l+1} \frac{\partial y_{l+1}}{\partial y_{l}} \frac{\partial y_{l}}{\partial w_{l}} - y_{l} \frac{\partial y_{l}}{\partial w_{l}} \\
\end{align}
Now we can simplify this equation by applying the chain rule:
\begin{equation}
\Delta w_{l} = -\frac{\partial C_{l+3}}{\partial w_{l}} - y_{l+2} \frac{\partial y_{l+2}}{\partial w_{l}} - y_{l+1} \frac{\partial y_{l+1}}{\partial w_{l}} - y_{l} \frac{\partial y_{l}}{\partial w_{l}}.
\end{equation}

The neural activities can be written in the quadratic form:
\begin{equation}
\Delta w_{l} = -\frac{\partial}{\partial w_{l}} \left( C_{l+3} + \frac{y_{l+2}^2}{2} + \frac{y_{l+1}^2}{2} + \frac{y_{l}^2}{2} \right).
\end{equation}
This equation can be generalized to any number of layers in the following manner:
\begin{equation}
\Delta w_{l} = -\frac{\partial}{\partial w_{l}} \left(C_{q} + \sum_{i=l}^{q-1} \frac{y_i^2}{2} \right)
\end{equation}

Where q is a upper layer with known target values, such as the output layer.

\end{document}